\pdfoutput=1

\documentclass[sigconf]{acmart}

\AtBeginDocument{%
  }

\copyrightyear{2023}
\acmYear{2023}
\setcopyright{acmlicensed}\acmConference[MM '23]{Proceedings of the 31st ACM International Conference on Multimedia}{October 29-November 3, 2023}{Ottawa, ON, Canada}
\acmBooktitle{Proceedings of the 31st ACM International Conference on Multimedia (MM '23), October 29-November 3, 2023, Ottawa, ON, Canada}
\acmPrice{15.00}
\acmDOI{10.1145/3581783.3612427}
\acmISBN{979-8-4007-0108-5/23/10}

\usepackage{multirow}
\usepackage{enumitem}

\begin{document}

\newcommand{\myet}{\emph{et~al.}}
\newcommand{\myeg}{\emph{e.g.}}
\newcommand{\myie}{\emph{i.e.}}
\newcommand{\myec}{\emph{etc}}
\newcommand{\by}[1]{\textcolor{magenta}{#1}}

\title{Unifying Two-Stream Encoders with Transformers for Cross-Modal Retrieval}

\author{Yi Bin}
\authornote{Both authors contributed equally to this research.}
\affiliation{%
  \institution{National University of Singapore}
  \country{Singapore}
}
\email{yi.bin@hotmail.com}

\author{Haoxuan Li}
\authornotemark[1]
\affiliation{%
  \institution{University of Electronic Science and Technology of China}
  \city{Chengdu}
  \country{China}
}
\email{lhx980610@gmail.com}

\author{Yahui Xu}
\affiliation{%
  \institution{University of Electronic Science and Technology of China}
  \city{Chengdu}
  \country{China}
}
\email{yahui2727@163.com}

\author{Xing Xu}
\affiliation{%
  \institution{University of Electronic Science and Technology of China}
  \city{Chengdu}
  \country{China}
}
\email{xing.xu@uestc.edu.cn}

\author{Yang Yang}
\authornote{Yang Yang is the corresponding author (Email: yang.yang@uestc.edu.cn).}
\affiliation{%
  \institution{University of Electronic Science and Technology of China}
  \city{Chengdu}
  \country{China}
}
\email{yang.yang@uestc.edu.cn}

\author{Heng Tao Shen}
\affiliation{%
  \institution{University of Electronic Science and Technology of China}
  \city{Chengdu}
  \country{China}
}
\email{shenhengtao@hotmail.com}


\renewcommand{\shortauthors}{Bin et al.}

\begin{abstract}
  Most existing cross-modal retrieval methods employ two-stream encoders with different architectures for images and texts, \myeg{}, CNN for images and RNN/Transformer for texts. Such discrepancy in architectures may induce different semantic distribution spaces and limit the interactions between images and texts, and further result in inferior alignment between images and texts. To fill this research gap, inspired by recent advances of Transformers in vision tasks, we propose to unify the encoder architectures with Transformers for both modalities. Specifically, we design a cross-modal retrieval framework purely based on two-stream Transformers, dubbed \textbf{Hierarchical Alignment Transformers (HAT)}, which consists of an image Transformer, a text Transformer, and a hierarchical alignment module. With such identical architectures, the encoders could produce representations with more similar characteristics for images and texts, and make the interactions and alignments between them much easier. Besides, to leverage the rich semantics, we devise a hierarchical alignment scheme to explore multi-level correspondences of different layers between images and texts. To evaluate the effectiveness of the proposed HAT, we conduct extensive experiments on two benchmark datasets, MSCOCO and Flickr30K. Experimental results demonstrate that HAT outperforms SOTA baselines by a large margin. Specifically, on two key tasks, \textit{i.e.}, image-to-text and text-to-image retrieval, HAT achieves 7.6\% and 16.7\% relative score improvement of Recall@1 on MSCOCO, and 4.4\% and 11.6\% on Flickr30k respectively. The code is available at \by{\url{https://github.com/LuminosityX/HAT}}.

\end{abstract}

\ccsdesc[500]{Networks~Network design principles}
\ccsdesc[300]{Information systems~Retrieval models and ranking}
\ccsdesc[100]{Computing methodologies~Information extraction}

\keywords{cross-modal retrieval, discrepancy in architectures, hierarchical alignment transformers.}

\maketitle

\section{Introduction}
\label{sec:intro}
Humans perceive and interact with the physical world via variant ways, \myeg{}, vision, sound and tactile sense. To make the machine simulate the perceiving process, simultaneously analyzing data from multiple modalities is a fundamental and important ability. Visual information and text data are the two most prevalent
modalities in our daily life, and the research of vision and language also has been attracting broad attention in the past couples of years~\cite{karpathy2015deep,yang2018video,bin2017adaptively,lee2018stacked,frome2013devise}. 
Benefiting from the great successes of deep learning in computer vision (CV) and natural language processing (NLP), many tasks and problems associating vision and language for multi-modal analysis also have achieved tremendous progress, such as cross-modal retrieval~\cite{shen2020exploiting,lee2018stacked,frome2013devise,wang2017adversarial,li2023your}, 
visual question answering (VQA)~\cite{peng2021progressive,antol2015vqa}, visual captioning~\cite{karpathy2015deep,bin2021multi}, and other tasks~\cite{chen2019neural}. In this paper, we focus on the study of cross-modal retrieval, 
which is a fundamental multi-modal understanding task and benefits plenty of multimedia applications. However, it is still very challenging for accurate retrieval, due to the requirements of exploring precise cross-modal alignment and comprehensive inter/intra-modal relations and interactions.

To align images and texts, early works apply canonical correlation analysis (CCA) to establish inter-modal associations between different modalities with subspace learning~\cite{klein2015associating,rasiwasia2010new,gong2014multi,pereira2013role}, or employ topic models to capture the relationship in the multi-modal joint distribution space~\cite{blei2003modeling,putthividhy2010topic,jia2011learning}. 
With the development of deep learning techniques, deep neural networks, \myeg{}, CNN and RNN, are applied to extract visual and textural representations, followed with projection functions to learn the mapping from the uni-modal to cross-modal spaces~\cite{frome2013devise,kiros2014unifying,wang2023quaternion}. 
However, such holistic uni-modal representations capture only the salient instances in the images or texts while ignore the non-salient ones or subtle relationships among instances.
It therefore fails to explore the comprehensive fine-grained semantic associations underlying images and texts. To tackle this problem, many works step further to devise fine-grained alignment frameworks~\cite{karpathy2015deep,niu2017hierarchical}, which first associate image regions and text words on the fragment-level, and aggregate the matched fragment pairs to obtain the final image-text pair. Besides, based on the fragments, rich intra-modal interactions also could be explored and used to improve the retrieval accuracy. 

The above approaches extract image and text features in two independent streams, which cannot incorporate any cross-modal interactions, with the heterogeneous gap between modalities remained. Actually, semantics in different modalities are complicated and diverse, which means that borrowing information from other modality may lead to better semantic representations. 
To this end, some works model cross-modal interactions to help with the uni-modal representation learning, for example, with the idea of fragment-level alignment~\cite{lee2018stacked,qu2021dynamic,wehrmann2020adaptive,chen2023inter}, which can be implemented by recurrent cross-modal messages passing, attentional aggregation, and \myec{}. 
From another perspective, motivated by the powerful representations of pre-trained Transformer in NLP, \myeg{}, BERT~\cite{kenton2019bert}, ALBERT~\cite{lan2019albert} and BART~\cite{lewis2020bart}. Many researchers dedicate to designing unified single-stream multi-modal representation pre-training frameworks ~\cite{li2021unimo,li2020oscar,xu2023multi}, and have achieved inspiring progress. However, comprehensive interactions across modalities within single-stream are computationally expensive and suffer from high latency comparing with two-stream methods during inference, because they need to extract the representations of the given image-text pair via the whole model. 

In this paper, we focus on the two-stream framework, which is computational-friendly and more applicable for real world scenarios. Compared with previous works~\cite{li2021unimo,li2020oscar} which apply different visual or textural backbones, \myeg{}, CNN and RNN/Transformer, we design an entire Transformer-based model, namely \textit{Hierarchical Alignment Transformers (HAT)}, to eliminate the architecture discrepancy between encoders in two-stream image-text matching framework.
Specifically, motivated by the recent successes of vision Transformers, \myeg{}, ViT~\cite{dosovitskiy2020image}, MAE~\cite{he2021masked}, and Swin Transformer~\cite{liu2021swin}, we employ Transformers for both image and text representation learning steams. The unified architectures for different modalities make the learned representations more compatible for semantic mapping and similarity measuring. Besides, previous works~\cite{zeiler2014visualizing,mahendran2016visualizing} point out that the shallow layers in CNN learn the texture features and the higher layers learn more semantic aspects. Transformers also exhibit similar property, \myeg{}, shallow layers in BERT tend to learn the static representation of entities while the higher ones capture more semantics in context~\cite{van2019does,de2020s}. Towards this end, we introduce a hierarchical alignment strategy to capture the rich correspondences between images and texts with different semantic levels. 
Most existing cross-modal retrieval methods only utilize single-level alignment to associate images and texts, while our proposed HAT integrates multi-level associations for final retrieval results.
In summary, the main contributions of this paper are:
\begin{itemize}[leftmargin=*]
    \item We propose to unify the two-stream encoders of images and texts with Transformers in cross-modal retrieval, which makes the learned image and text representations more compatible for aligning images and texts.
    \item To comprehensively explore cross-modal interactions, we propose a hierarchical alignment strategy associating images and texts with multi-level semantic clues.
    \item Extensive experiments have been conducted on two commonly-used datasets, \myie{}, MSCOCO and Flickr30K. The experimental results demonstrate that our proposed HAT outperforms all the SOTAs by a large margin.
\end{itemize}

\section{Related Works}
\subsection{Cross-Modal Retrieval}

\textbf{Global matching} aims to explore the correlation by projecting the entire image and text into a common semantic space.
The early method proposed by Frome \textit{et al.}~\cite{frome2013devise}, utilizing a CNN and Skip-Gram~\cite{mikolov2013efficient} for visual and language feature extraction, respectively.
Kiros \textit{et al.}~\cite{kiros2014unifying} made the first attempt to encode text using a GRU and designed a triplet ranking loss to optimize the model.
To make more usage of informative pairs, some researchers have turned to design a more novel objective function. Faghri \textit{et al.}~\cite{faghri2017vse++}integrated the hard negative mining technology into the triplet loss function. 
Recently, Zheng \textit{et al.}~\cite{zheng2020dual} proposed a discriminative feature embedding method with instance loss. Although these methods are very effective, they ignore the local cues between regions and words.

\textbf{Local matching } steps further to explore fine-grained correspondences between images and texts by extracting local features (\textit{e.g.}, image regions and text words). Karpathy \textit{et al.}~\cite{karpathy2015deep} first proposed to extract features for each image region 
with R-CNN~\cite{girshick2014rich}, and then used the most similar region-word pairs for image-text matching. With the great success of attention mechanism~\cite{bahdanau2014neural}, Niu \textit{et al.}~\cite{niu2017hierarchical} extracted phrase-level features of text by using the LSTM to further align image regions with text words. 
Following the attention mechanism, a stacked cross attention model was proposed by Lee~\textit{et al.}~\cite{lee2018stacked}, which finds all latent alignments between them by selectively aggregating regions and words.
Later, Wu \textit{et al.}~\cite{wu2019learning} introduced the self-attention mechanism and Qu \textit{et al.}~\cite{qu2021dynamic} designed four types of interaction cells for cross-modal interaction, and introduced a dynamic routing mechanism to dynamically select interaction paths based on inputs.
More recently, Zhang \textit{et al.}~\cite{zhang2022negative} proposed Negative-Aware Attention Framework based on the SCAN method, which focuses on mining the negative effects of mismatched fragments and ultimately achieves more accurate similarity calculation. Chen \textit{et al.} claimed that the strategy for feature aggregation is crucial, and thus proposed the VSE$\infty$~\cite{chen2021learning} method, which learns the most suitable and best pooling strategy via a generalized pooling operator (GPO)and sets a new SOTA.

\begin{figure*}
\begin{center}
\includegraphics[width=0.95\textwidth]{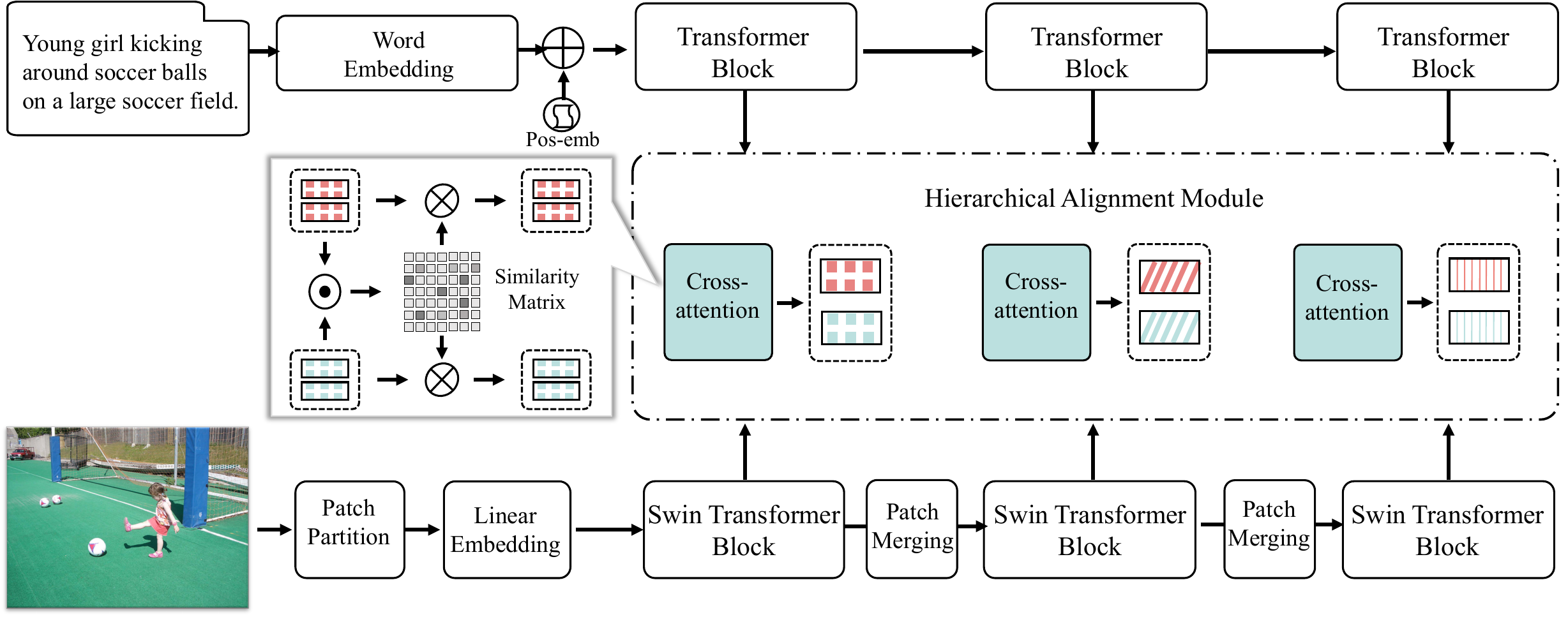}
\end{center}
\caption{
The overall flowchart of the proposed HAT. HAT consists of two-stream of Transformers, BERT and Swin Transformer in specific, for image and text encoding, and a hierarchical alignment module equipped with cross-attention mechanism. The hierarchical alignment module aligns the image and text representations from different semantic levels.
}
\label{fig:overview}
\end{figure*}

\subsection{Transformers}
\textbf{Language Transformers} is first proposed by Vaswani et al.~\cite{vaswani2017attention} for machine translation, which is solely based on attention mechanisms, \myie{}, self-attention and multi-head attention. In the past few years, it has been widely used in natural language processing, \myeg{}, word representation, sentence representation and question answering, and achieved the state of the arts on a wide range of tasks. Motivated by the great success of the pre-training and fine-tuning scheme in computer vision, Howard~\myet{}~\cite{howard2018universal} proposed to design and pre-train a universal language model, based on Transformer, for text classification, which significantly outperforms the state of the arts and encourages the community to explore the potentials of pre-training for other NLP tasks. BERT~\cite{kenton2019bert} is one of the most famous and popular works, which employs several Transformer blocks for sentence encoding and designs a masked language model to learn the comprehensive and contextual word representation. The simple yet powerful BERT could be pre-trained by self-supervision with very large amount of data, yielding gorgeous results on a various of NLP tasks. This makes the community step further to design powerful architectures with Transformer to understand the language by mining large scale data. Brown~\myet{}~\cite{brown2020language} devised an extremely large Generative Pre-trained Transformer $3,175$ billion parameters, as known as GPT-3, pre-trained on 45TB plain text data, which could achieve strong performance on various downstream tasks without any fine-tuning. Besides, XLNET~\cite{yang2019xlnet}, ALBERT~\cite{lan2019albert}, and T5~\cite{raffel2020exploring} are some representative works of large scale pre-training Transformers. These powerful Transformer-based models, with the strong representation capacity, have achieved significant breakthroughs in every respect of NLP.

\textbf{Vision Transformers} is by the great progress and success of Transformer architectures in the field of NLP, and have been applied to replace CNNs on computer vision tasks. Chen~\myet{}~\cite{chen2020generative} proposed a pioneering generative pre-training Transformer framework that predicts pixels instead language tokens to learn visual representation with self-supervision. Their pixel Transformer treats the images as pixel sequences and auto-regressively predicts the pixels without 2D structure, and still achieves competitive results on several classification tasks. ViT~\cite{dosovitskiy2020image} directly implemented a pure Transformer to the sequences of image patches and performs well on image classification with sufficient training data. Another popular work is Swin-Transformer~\cite{liu2021swin}, constructing hierarchical feature maps via a novel shifted windowing scheme and retaining only linear computational complexity to image size. With such hierarchical feature maps, the Swin Transformer could be conveniently equipped with advanced techniques and applied to variants downstream applications, \myeg{}, object detection, semantic and segmentation. More recently, He~\myet{}~\cite{he2021masked} devised a simple framework of masked autoencoders (MAE) with an asymmetric encoder-decoder architecture to operate only on the visible subset of patches and reconstruct the original image from the latent representation and masked tokens. The MAE also reveals that masking a high proportion of the input image could yield nontrivial results, as well as accelerating the training procedure. In addition to above works, there also exist many vision Transformers for other vision tasks, including video understanding~\cite{zhou2018end,wang2021bevt}, instance/semantic segmentation~\cite{wang2021end,xie2021segformer}, and multi-modal understanding~\cite{li2020oscar,li2021unimo}.

\section{The Proposed Approach}
Most existing cross-modal retrieval approaches~\cite{lee2018stacked,qu2021dynamic,wang2023multilateral} employ different architectures for representation learning, \myeg{}, CNNs as image encoder and RNNs/Transformers as text encoder, and measure the similarity between images and texts only utilizing single level correspondences. 
The discrepancy between the encoder structures of images and texts leads to different feature property and may result in inferior cross-modal performance.
In this works, we propose a novel framework, termed as Hierarchical Alignment Transformers (HAT), to unify the two-stream encoder architectures with Transformers~\cite{vaswani2017attention}, and hierarchically align images and texts with multi-level semantic correspondences. As the visual pipeline illustrated in Figure~\ref{fig:overview}, the proposed HAT consists of Text Transformer, Image Transformer and Hierarchical Alignment Modules. 

\subsection{Text Transformer}
Given a text sentence $S=\{w_t|w_1, w_2,...,w_T\}$ consisting of $T$ words, the basic idea is to employ a recurrent neural network, \myeg{}, an LSTM or GRU, to learn the sentence representation. To capture fine-grained correspondences between images and texts, the output state of each step also be utilized as the local word representation~\cite{lee2018stacked}. To enable such fine-grained association, we employ a Transformer structure, BERT in specific, to extract the contextual representations of words, which could easily explore the bidirectional context interactions between all the word pairs.
Specifically, given a sentence $S$, BERT first projects each word $w_t$ into a contiguous embedding space $S_e = \{e_t|e_1, e_2,...,e_T\}$.
Then, each word embedding $e_t$ is integrated with a positional embedding and a segment embedding via addition to feed into the Transformer blocks, and learn the contextual representation of each word with attention.
We finally obtain the word-level representations for the sentence $S$, denoted as $S^l = \{w^l_t| w_1^l, w_2^l,...,w_T^l \}$, where $l$ denotes the $l$-th Transformer block of BERT. In this work, we use BERT-base consisting of 12 Transformer blocks. As pointed out in~\cite{van2019does,de2020s}, different layers in BERT capture different levels of semantics, we therefore adopt the outputs of multiple Transformer blocks for multi-level semantic representations of sentences and compute the hierarchical associations with image representations.

\subsection{Image Transformer}
Different from adopting CNN-based models to learn image features, \myeg{}, faster-RCNN~\cite{ren2015faster} or ResNet~\cite{he2016deep}, inspired by the recent advances of vision Transformers~\cite{he2021masked,liu2021swin}, we employ Transformer-based model to encapsulate the images for dense representations and unify the backbones of different modalities. Specifically, we implement the recent successful Swin Transformer~\cite{liu2021swin} as our vision backbone, which significantly outperforms CNN and shows great potential in many vision tasks. Swin Transformer, equipped with a shifted windowing scheme, is able to construct hierarchical feature maps and has linear computational complexity to image size.

Given an image $I\in \mathbb{R}^{3\times H\times W}$, we first split the image into several non-overlapping patches via the patch partition module, as most existing vision Transformers. Each patch is with the size of $4\times4$ pixels and treated as a ``token'' to feed into the Transformer blocks. Inspired by previous work~\cite{chen2020image}, we take the output feature maps of multiple stages as hierarchical representations of images to express more rich semantics, and denote the output feature map of each stage as $I^i$, where $i$ indicates the $i$-th stage. In specific, we discard the output of ``Stage 1'' due to its large number of tokens that results in high dimension and expensive computation costs. The outputs of subsequent stages, \myie{}, Stage 2-4, are also termed as low-, middle-, and high-level semantic representations of images, then the overall image hierarchical representation $I_h$ could be denoted as:
\begin{equation}
    I_h = SwinT(I) = \{ I^{s2}, I^{s3}, I^{s4}\}.
\end{equation}
To be consistent with BERT layer representation, we convert stages in $I^{s2}$, $I^{s3}$, and $I^{s4}$ to corresponding layers, \myie{}, 4, 10, 12-th layer in Swin Transformer.

\subsection{Hierarchical Alignment Module}
After obtained the representations of images and texts, the important thing is to align them with semantic correspondences. Most existing works measure the similarities between images and texts with the last layers of two-stream encoders~\cite{lee2018stacked}. As previous research pointed out, different layers of BERT capture different-level semantics~\cite{van2019does,de2020s}, \myeg{}, shallow layers tend to convey more static and local semantics and the deep layers express more contextual information. This observation is also consistent with the visual representation captured by CNNs~\cite{zeiler2014visualizing,mahendran2016visualizing}. Therefore, to enable our proposed model align images and texts leveraging rich semantics, we design a hierarchical alignment scheme that simultaneously computes the similarity between given image-text pair with the representations of multiple levels. As the cross-attention modules illustrated in the middle of Figure~\ref{fig:overview}, our hierarchical alignment integrates three-level semantics, \myie{}, low-, middle-, and high-level, to learn the associations between images and texts. Swin Transformer consists of four stages of modules, divided by the patch partition operations. We simply adopt these stages as our multi-level semantics split, excepting ``Stage 1'' due to its large number of tokens that would be computationally expensive. For hierarchical representations of text Transformer, because BERT has the same number of layers, \myie{}, 12 layers, as Swin Transformer, we symmetrically leverage the outputs of the same layers of BERT for multi-level semantic representations. In other words, the the outputs of 4-, 10-, and 12-th layers of both Swin Transformer and BERT are extracted for our overall representations of images and texts, respectively. We compute the cosine similarities for each level and combine the hierarchical similarities by addition.

For the cross-modal alignment of each level, previous works~\cite{lee2018stacked,xu2020cross,qu2021dynamic} have verified that fine-grained alignment with local semantic aspects, \myeg{}, visual regions and words, could significantly improve the cross-modal retrieval performance. Towards this end, we compute the similarity between a pair of given image and text by aligning the representations of ``region tokens'' and words, outputted by Swin Transformer and BERT. Specifically, inspired by the effectiveness of the stacked cross attention mechanism, we implement image-text and text-image stacked cross attention respectively. Supposing the output of $l$-th layer ($l$ could be 4, 10, 12 in this work) of Swin Transformer $I^l=\{v_k^l|v_1^l,v_2^l,...,v_K^l\}$ with $K$ ``region tokens'', and the output of $l$-th layer of BERT $S^l = \{w^l_t| w_1^l, w_2^l,...,w_T^l \}$ with $T$ words, the \textbf{image-text} stacked cross attention first computes the similarities between all region-word pairs:
\begin{equation}
    s_{kt}^l = \frac{{v_k^l}^T w_t^l}{||v_k^l||||w_t^l||},
\end{equation}
where $s_{kt}^l$ denotes the similarity between $k$-th region and $t$-th word in $l$-th layer of both Swin Transformer and BERT. Based on the region-word similarity, we aggregate the weighted words representations for each image region as:
\begin{equation}
    a_k^l = \sum^{T}_{t=1}\alpha_{kt}^lw_t^l,
\end{equation}
\begin{equation}
    \alpha_{kt}^l = \frac{exp(\lambda * \bar{s}_{kt}^l)}{\sum^T_{t=1}exp(\lambda * \bar{s}_{kt}^l)},
\end{equation}
where $\lambda$ denotes the temperature, and $\bar{s}_{kt}^l$ is normalized similarity:
\begin{equation}
\bar{s}_{kt}^l=\frac{[s_{kt}^l]_+}{\sqrt{\sum_{k=1}^K[s_{kt}^l]^2_+}},
\end{equation}
where $[x]_+=max(x,0)$ denotes the hinge function to keep the similarity with positive value. Finally, we measure the cosine similarity between each region-word pair $(v_k^l,a_k^l)$ based on image-text stacked cross attention, and aggregate all the pairs for the overall similarity between given image and text pair.
The \textbf{text-image} stacked cross attention follows the similar formulations.

With such hierarchical alignment, the proposed model is able to capture multi-level semantic correspondences for image-text matching, and successfully achieves significant improvement in the evaluation of all the metrics.

\section{Experiments}
To verify the effectiveness of our proposed hierarchical alignment Transformers, we conduct extensive experiments on two commonly used datasets, namely Flickr30k and MSCOCO. We mainly examine the cross-modal retrieval under two dual scenarios: Image-to-Text retrieval (I2T) and Text-to-Image retrieval (T2I). 

\subsection{Datasets and Evaluation Metrics}

\textbf{Flickr30K}~\cite{young2014image} consists of 31,783 images collected from Flickr, each of which is annotated with 5 description sentences. Following previous works~\cite{karpathy2015deep, lee2018stacked, liu2019focus}, we split this dataset into 29,783, 1,000, 1,000 iamges for training, validation, and testing respectively. {\bf MSCOCO}~\cite{chen2015microsoft} contains 123,287 images, where each image is also described by 5 different sentences. For fair comparison, we follow previous work~\cite{lee2018stacked, qu2020context} using Karpathy split~\cite{karpathy2015deep}. 
We conduct two evaluation settings on \textbf{MSCOCO}, which are 1) \textbf{MSCOCO(5K)}: retrieving images within full 5K images or sentences within its corresponding corpus; 2) \textbf{MSCOCO(1K)}: 5-folds evaluation with each fold consisting of 1K images, and the final result is reported by averaging folds.

{\bf Evaluation Metrics} 
We follow previous works~\cite{lee2018stacked, liu2019focus} to evaluate the performance with the Recall$@$K metric, short in R@K (K=1, 5, 10). 
It measures the percentage of ground-truth hits in the top-K ranking list. The higher R@K indicates the better performance. 

\subsection{Implement Details}
We implement our HAT with PyTorch and optimize it by the Adam optimizer~\cite{kingma2014adam}. For the learning rate, we use a small learning rate starting with 1$e$-5 and decay the learning rate by dividing 10 for every 10 epochs. We employ the base Swin Transformer to extract the image features and BERT-base to extract text features. During the training , we first freeze the pre-trained models, \myie{}, Swin Transformers and BERT, for 10 epochs. This strategy avoids that the random initialization of hierarchical alignment module misleads the pre-trained weights to be optimized to inferior areas, and makes the hierarchical alignment module more compatible with the pre-trained parameters. Then the whole HAT framework is further fine-tune to more optimal scenario.
The dimensions of the visual text features extracted by the Swin Transformer and the BERT are 1024 and 768, respectively. Moreover, we implement the triplet loss, a variant for cross-modal retrieval in specific, as:
\begin{equation}
\label{equ:triplet_final}
\begin{aligned}
        \mathbf{\mathcal{L}} = &[m - sim(I, S) + sim(I, \bar{S})]_+ \\ + \ &[m - sim(I, S) + sim(\bar{I}, S)]_+ ,
\end{aligned}
\end{equation}
where $m$ is a margin set as 0.2, and $[\cdot]_+=max(0,\cdot)$. $\bar{S}$ and $\bar{I}$ denote the negative sample to push away. $sim(\cdot)$ measures the similarity between images and texts, and cosine similarity is used here:
\begin{equation}
    sim(I, S) = \sum_{l} \sum_{k} \frac{{v_k^l}^T a_k^l}{||v_k^l||||a_k^l||}.
\end{equation}

\begin{table*}
\centering
\caption{
Performance comparison between our proposed HAT and recent state of the arts on MSCOCO. MSCOCO(5K) and MSCOCO(1K) denote the evaluation settings of the full 5K and average of 5-folds 1K test images. For models with cross-attention, we exhibit the best single model and the ensemble model reported in the corresponding paper, while report all the single models, including i-t and t-i model of HAT for comprehensive comparison. We use ``single'' in bracket and superscript ``*'' to indicate the best single and ensemble models. The best results are in bold, and the best results of baselines are underlined.
} 
\setlength{\tabcolsep}{2.5mm}{
\begin{tabular}{c|cccccc|cccccc}
\hline
\multirow{3}{*}{Method} & \multicolumn{6}{c|}{MSCOCO(1K)}                                              & \multicolumn{6}{c}{MSCOCO(5K)}                                              \\ \cline{2-13} 
                        & \multicolumn{3}{c|}{Image-to-Text}      & \multicolumn{3}{c|}{Text-to-Image} & \multicolumn{3}{c|}{Image-to-Text}      & \multicolumn{3}{c}{Text-to-Image} \\ \cline{2-13} 
                        & R@1  & R@5  & \multicolumn{1}{c|}{R@10} & R@1        & R@5       & R@10      & R@1  & R@5  & \multicolumn{1}{c|}{R@10} & R@1       & R@5       & R@10      \\ \hline
SCAN(single)~\cite{lee2018stacked}            & 70.9 & 94.5 & \multicolumn{1}{c|}{97.8} & 56.4       & 87.0      & 93.9      & 46.4  & 77.4  & \multicolumn{1}{c|}{87.2}  & 34.4       & 63.7       & 75.7       \\
SCAN$^*$~\cite{lee2018stacked}                    & 72.7 & 94.8 & \multicolumn{1}{c|}{98.4} & 58.8       & 88.4      & 94.8      & 50.4 & 82.2 & \multicolumn{1}{c|}{90.0} & 38.6      & 69.3      & 80.4      \\
CAMP~\cite{wang2019camp}                    & 72.3 & 94.8 & \multicolumn{1}{c|}{98.3} & 58.5       & 87.9      & 95.0      & 50.1 & 82.1 & \multicolumn{1}{c|}{89.7} & 39.0      & 68.9      & 80.2      \\
BFAN(single)~\cite{liu2019focus}            & 73.7 & 94.9 & \multicolumn{1}{c|}{-}    & 58.3       & 88.4      & -         & -    & -    & \multicolumn{1}{c|}{-}    & -         & -         & -         \\
BFAN$^*$~\cite{liu2019focus}                    & 74.9 & 95.2 & \multicolumn{1}{c|}{-}    & 59.4       & 88.4      & -         & -    & -    & \multicolumn{1}{c|}{-}    & -         & -         & -         \\
SAEM~\cite{wu2019learning}                    & 71.2 & 94.1 & \multicolumn{1}{c|}{97.7} & 57.8       & 88.6      & 94.9      & -    & -    & \multicolumn{1}{c|}{-}    & -         & -         & -         \\
CAAN~\cite{zhang2020context}                    & 75.5 & 95.4 & \multicolumn{1}{c|}{98.5} & 61.3       & 89.7      & 95.2      & 52.5 & 83.3 & \multicolumn{1}{c|}{90.9} & 41.2      & 70.3      & 82.9      \\
DP-RNN~\cite{chen2020expressing}                  & 75.3 & 95.8 & \multicolumn{1}{c|}{98.6} & 62.5       & 89.7      & 95.1      & -    & -    & \multicolumn{1}{c|}{-}    & -         & -         & -         \\
VSRN~\cite{li2019visual}                    & 76.2 & 94.8 & \multicolumn{1}{c|}{98.2} & 62.8       & 89.7      & 95.1      & 53.0 & 81.1 & \multicolumn{1}{c|}{89.4} & 40.5      & 70.6      & 81.1      \\
SGM~\cite{wang2020cross}                     & 73.4 & 93.8 & \multicolumn{1}{c|}{97.8}    & 57.5       & 87.3      & 94.3         & 50.0 & 79.3 & \multicolumn{1}{c|}{87.9} & 35.3      & 64.9      & 76.5      \\
IMRAM~\cite{chen2020imram}                   & 76.7 & 95.6 & \multicolumn{1}{c|}{98.5} & 61.7       & 89.1      & 95.0      & 53.7 & 83.2 & \multicolumn{1}{c|}{91.0} & 39.6      & 69.1      & 79.8      \\
MMCA~\cite{wei2020multi}                    & 74.8 & 95.6 & \multicolumn{1}{c|}{97.7} & 61.6       & 89.8      & 95.2      & 54.0 & 82.5 & \multicolumn{1}{c|}{90.7} & 38.7      & 69.7      & 80.8      \\
GSMN(single)~\cite{liu2020graph}            & 76.1 & 95.6 & \multicolumn{1}{c|}{98.3} & 60.4       & 88.7      & 95.0      & -    & -    & \multicolumn{1}{c|}{-}    & -         & -         & -         \\
GSMN$^*$~\cite{liu2020graph}                    & 78.4 & 96.4 & \multicolumn{1}{c|}{98.6} & 63.3       & 90.1      & 95.7      & -    & -    & \multicolumn{1}{c|}{-}    & -         & -         & -         \\
ADAPT(single)~\cite{wehrmann2020adaptive}           & 75.3 & 95.1 & \multicolumn{1}{c|}{98.4} & 63.3       & 90.0      & 95.5      & -    & -    & \multicolumn{1}{c|}{-}    & -         & -         & -         \\
ADAPT$^*$~\cite{wehrmann2020adaptive}                   & 76.5 & 95.6 & \multicolumn{1}{c|}{\underline{98.9}} & 62.2       & 90.5      & 96.0      & -    & -    & \multicolumn{1}{c|}{-}    & -         & -         & -         \\
CAMERA(single)~\cite{qu2020context}          & 75.9 & 95.5 & \multicolumn{1}{c|}{98.6} & 62.3       & 90.1      & 95.2      & 53.1  & 81.3  & \multicolumn{1}{c|}{89.8}  & 39.0       & 70.5       & 81.5       \\
CAMERA$^*$~\cite{qu2020context}                  & 77.5 & 96.3 & \multicolumn{1}{c|}{98.8} & 63.4       & 90.9      & 95.8      & 55.1 & 82.9 & \multicolumn{1}{c|}{91.2} & 40.5      & 71.7      & 82.5      \\
DIME(single)~\cite{qu2021dynamic}            & 77.9 & 95.9 & \multicolumn{1}{c|}{98.3} & 63.0       & 90.5      & 96.2      & 56.1 & 83.2 & \multicolumn{1}{c|}{91.1} & 40.2      & 70.7      & 81.4      \\
DIME$^*$~\cite{qu2021dynamic}                    & 78.8 & 96.3 & \multicolumn{1}{c|}{98.7} & \underline{64.8}       & \underline{91.5}      & \underline{96.5}      & \underline{59.3} & \underline{85.4} & \multicolumn{1}{c|}{91.9} & \underline{43.1}      & \underline{73.0}      & 83.1      \\
VSE$\infty$~\cite{chen2021learning}                    & 79.7 & 96.4 & \multicolumn{1}{c|}{\underline{98.9}} & \underline{64.8}       & 91.4      & 96.3      & 58.3 & 85.3 & \multicolumn{1}{c|}{\underline{92.3}} & 42.4      & 72.7      & \underline{83.2}   \\ 
NAAF~\cite{zhang2022negative}                    & \underline{80.5} & \underline{96.5} & \multicolumn{1}{c|}{98.8} & 64.1       & 90.7      & \underline{96.5}      & 58.9 & 85.2 & \multicolumn{1}{c|}{92.0} & 42.5      & 70.9      & 81.4   \\ 
\hline
HAT(i-t)                & 81.3  & 97.0  & \multicolumn{1}{c|}{99.2}  & 68.9        & 92.7       & 97.1       & 61.6  & 87.3  & \multicolumn{1}{c|}{93.4}  & 47.9       & 76.4       & 85.8       \\
HAT(t-i)                & {81.8}  & 96.7  & \multicolumn{1}{c|}{99.1}  & 69.8        & 93.0       & 97.1       & 61.9  & {87.6}  & \multicolumn{1}{c|}{93.3}  & 48.8       & 77.3       & 86.1       \\
HAT$^*$                     & \textbf{82.6}  & \textbf{97.4}  & \multicolumn{1}{c|}{\textbf{99.3}}  & \textbf{70.8}        & \textbf{93.3}       & \textbf{97.4}       & \textbf{63.8}  & \textbf{88.5}  & \multicolumn{1}{c|}{\textbf{94.1}}  & \textbf{50.3}       & \textbf{78.2}       & \textbf{86.9}       \\ \hline
\end{tabular}
} 
\label{tab:mscoco}
\end{table*}

\subsection{Comparison Baselines}
\label{sec:baseline}
To evaluate the effectiveness of proposed hierarchical alignment Transformers, we compare the cross-modal retrieval performance with several state of the art methods. As mentioned above, existing methods could be grouped into several paradigms based on the involved interaction type.

\noindent\textbf{Intra-modal interaction:} Methods that focus on intra-modal interactions, such as SAEM~\cite{wu2019learning}, VSRN~\cite{li2019visual}, CAMERA~\cite{qu2020context}, and VSE$\infty$~\cite{chen2021learning}. SAEM~\cite{wu2019learning} models images and text separately through a self-attention mechanism. VSRN~\cite{li2019visual} performs semantic reasoning by constructing region graphs in order to employ graph convolutional networks. CAMERA~\cite{qu2020context} then further designs a gated self-attention mechanism for context multi-view modeling. VSE$\infty$~\cite{chen2021learning} proposes GPO to generate the best pooling strategy to improve the semantic representation learning.

\noindent\textbf{Inter-modal interaction:} Methods that interact through different cross-modal operations, including SCAN~\cite{lee2018stacked}, CAMP~\cite{wang2019camp}, BFAN~\cite{liu2019focus}, IMRAM~\cite{chen2020imram}, ADAPT~\cite{wehrmann2020adaptive}, and NAAF~\cite{zhang2022negative}. SCAN~\cite{lee2018stacked} and BFAN~\cite{liu2019focus} selectively focus on text-related regions and image-related words via cross-modal attention. CAMP~\cite{wang2019camp} combines image and text for modal interaction via cross-modal gated fusion operations. IMRAM~\cite{chen2020imram} tries to mine deeper levels of modal interactions by iterative way. NAAF~\cite{zhang2022negative} refines similarity calculation by considering the negative impact caused by mismatched fragments.

\noindent\textbf{Hybrid interaction:} Methods that simultaneously perform inter- and intra-modal interactions are CAAN~\cite{zhang2020context}, DP-RNN~\cite{chen2020expressing}, SGM~\cite{wang2020cross}, MMCA~\cite{wei2020multi}, GSMN~\cite{liu2020graph}, and DIME~\cite{qu2021dynamic}. Hybrid interaction leads to better semantic alignment and retrieval performance.

Several methods implement cross attention mechanism with different directions, \myeg{}, image-text (i-t) and text-image (t-i) cross attention, and ensemble models of two directions, including SCAN, BFAN, CAMP, and ADAPT, as well as our HAT. We present both single and ensemble models for comprehensive comparison. As listed in Table~\ref{tab:flickr}, models with ``*'' denote the ensemble ones. For single models, due to the space limitation, we only list the best one of i-t and t-i directions, reported in each paper, and add ``single'' in the bracket to indicate this. For our HAT, we illustrate single models with cross-modal guidance of both i-t and t-i directions, namely HAT(i-t) and HAT (t-i), as well as the full ensemble HAT$^*$. 

\begin{table}[]
\centering
\caption{
Comparison results on the Flick30K. Symbols are the same as the ones in Table~\ref{tab:mscoco}.
} 
\setlength{\tabcolsep}{1.2mm}{
\begin{tabular}{c|ccc|ccc}
\hline
\multirow{2}{*}{Method} & \multicolumn{3}{c|}{Image-to-Text}            & \multicolumn{3}{c}{Text-to-Image}             \\ \cline{2-7} 
                        & R@1           & R@5           & R@10          & R@1           & R@5           & R@10          \\ \hline
SCAN(single)~\cite{lee2018stacked}            & 67.9          & 89.0          & 94.4          & 43.9          & 74.2          & 82.8          \\
SCAN$^*$~\cite{lee2018stacked}                    & 67.4          & 90.3          & 95.8          & 48.6          & 77.7          & 85.2          \\
CAMP~\cite{wang2019camp}                    & 68.1          & 89.7          & 95.2          & 51.5          & 77.1          & 85.2          \\
BFAN(single)~\cite{liu2019focus}            & 65.5          & 89.4          & -             & 47.9          & 77.6          & -             \\
BFAN$^*$~\cite{liu2019focus}                    & 68.1          & 91.4          & -             & 50.8          & 78.4          & -             \\
SAEM~\cite{wu2019learning}                    & 69.1          & 91.0          & 95.1          & 52.4          & 81.1          & 88.1          \\
CAAN~\cite{zhang2020context}                    & 70.1          & 91.6          & 97.2          & 52.8          & 79.0          & 87.9          \\
DP-RNN~\cite{chen2020expressing}                  & 70.2          & 91.6          & 95.8          & 55.5          & 81.3          & 88.2          \\
VSRN~\cite{li2019visual}                    & 71.3          & 90.6          & 96.0          & 54.7          & 81.8          & 88.2          \\
SGM~\cite{wang2020cross}                     & 71.8          & 91.7          & 95.5          & 53.5          & 79.6          & 86.5          \\
IMRAM~\cite{chen2020imram}                   & 74.1          & 93.0          & 96.6          & 53.9          & 79.4          & 87.2          \\
MMCA~\cite{wei2020multi}                    & 74.2          & 92.8          & 96.4          & 54.8          & 81.4          & 87.8          \\
GSMN(single)~\cite{liu2020graph}            & 72.6          & 93.5          & 96.8          & 53.7          & 80.0          & 87.0          \\
GSMN$^*$~\cite{liu2020graph}                    & 76.4          & 94.3          & 97.3          & 57.4          & 82.3          & 89.0          \\
ADAPT(single)~\cite{wehrmann2020adaptive}           & 73.6          & 93.7          & 96.7          & 57.0          & 83.6          & 90.3          \\
ADAPT$^*$~\cite{wehrmann2020adaptive}                   & 76.6           & 95.4          & 97.6          & 60.7          & 86.6          & 92.0          \\
CAMERA(single)~\cite{qu2020context}          & 76.5          & 95.1          & 97.2          & 58.9          & 84.7          & 90.2          \\
CAMERA$^*$~\cite{qu2020context}                  & 78.0          & 95.1          & 97.9          & 60.3          & 85.9          & 91.7          \\
DIME(single)~\cite{qu2021dynamic}            & 77.4          & 95.0          & 97.4          & 60.1          & 85.5          & 91.8          \\
DIME$^*$~\cite{qu2021dynamic}                    & 81.0          & 95.9          & \underline{98.4}          & \underline{63.6}          & \underline{88.1}          & \underline{93.0}          \\ 
VSE$\infty$~\cite{chen2021learning} & 81.7          & 95.4          & 97.6          & 61.4          & 85.9          & 91.5          \\ 
NAAF~\cite{zhang2022negative} & \underline{81.9}          & \underline{96.1}          & 98.3          & 61.0          & 85.3          & 90.6          \\
\hline
HAT(i-t)                & 83.9          & 96.7          & 98.3          & 68.8          & 90.5          & 94.6          \\
HAT(t-i)                & 84.3          & {97.3} & {98.9} & 70.2          & 90.8          & 94.5          \\
HAT$^*$                     & \textbf{85.5} & \textbf{97.3}          & \textbf{99.0}          & \textbf{71.0} & \textbf{91.2} & \textbf{95.1} \\ \hline
\end{tabular}
} 

\label{tab:flickr}
\end{table}

\subsection{Performance Comparison}
The overall comparisons with state of the arts for MSCOCO and Flickr30K are exhibited in Table~\ref{tab:mscoco} and Table~\ref{tab:flickr}, respectively. From the result comparison, we have several observations as follows:

\begin{itemize}[leftmargin=*]

    \item Our proposed HAT$^*$ outperforms all the baselines with significant improvements in all metrics, for both MSCOCO and Flickr30K benchmarks, and all the single HAT models, namely HAT(i-t) and HAT(t-i) for different guidance directions, show the similar results. Comparing with the state of the art, our best results (in bold) gain 7.6\% and 16.7\% relative score improvements for text retrieval and image retrieval, in Recall@1 on MSCOCO (5K). For Flickr30K, our proposed approach improves the corresponding performance of text retrieval and image retrieval by 4.4\% and 11.6\% (Recall@1) relatively. 
    We also note that our HAT boosts the performance of text-to-image retrieval with much larger gap (16.7\% \textbf{v.s.} 7.6\%, and 11.6\% \textbf{v.s.} 4.4\%) than image-to-text retrieval. This phenomenon mainly comes from that the overall performance of text-to-image retrieval is inferior comparing with image-to-text retrieval, resulting in a smaller value, which has more scope to improve. 
    Overall, these observations show the superiority of our proposed HAT for cross-modal retrieval.
    
    \item As described in Section~\ref{sec:baseline}, we can divide the baselines into three groups based on the type of modality interaction. Obviously, benefiting from the rich interaction between images and texts, most works adopt inter-modal interaction or hybrid interaction (also involving inter-modal) to learn the semantic associations between them, and usually achieve better retrieval performances. Inter-modal interaction attempts to enrich the representations of images (or texts) with the counterpart guidance. With such cross-modal interactions, the models tend to learn the representation involving heterogeneous information of both modalities and narrow the semantic gap between representations of images and texts, which is similar to the cross-modal pre-training~\cite{li2020oscar,li2021unimo} and leads to better performance. However, CAMERA only exploring the intra-modal interaction still outperforms most approaches and results in the second competitive baseline. The competitive performance of CAMERA mainly attributes to the design of adaptive gating module to capture contextual information and the leverage of multi-view summarization, which indicates that comprehensive and elaborate intra-modal interactions exploration are also essential and crucial for boosting the performance of the cross-modal retrieval.
    
    \item It is well known that powerful feature representations usually lead to impressive performance in many tasks~\cite{kenton2019bert,he2016deep,he2021masked}, but have not been well discussed in previous cross-modal retrieval works. We note that almost all the baselines leverage Faster-RCNN or Bottom-Up as image feature extractor, and (Bi-)RNN or BERT as text representation encoder. As a powerful pre-trained language model, BERT and its successors demonstrate superiority in almost all the NLP tasks, including language understanding, generation, and infilling.
    Similar phenomenons are observed in this work, in other words, the methods adopting BERT as text encoder, \myeg{}, DIME, NAAF, RVSE++, and CAMERA, achieve more precise retrieval results. In the visual part, different from all the baselines, our HAT implements the recent powerful model, Swin Transformer, to extract visual representations and result in significant improvement. But we want to claim that the remarkable improvement of HAT not only comes from the powerful pre-trained model, but also benefits from the unified structures of image and text encoders, \myie{}, Transformers, which will be comprehensively investigated and discussed in Section~\ref{sec:disc_encoder}.

\end{itemize}

\subsection{Investigation of Structural Discrepancy between Image and Text Encoders}
As claimed above, using different structures of encoders for images and texts may degrade the representation learning of each modality and also affect the semantic alignment across modalities. In this section, we will empirically investigate whether the discrepancy of encoders in structure has impact on the cross-modal retrieval task and how it affects the final performance. Towards this end, we test different encoder combos, combining image and text encoders with different architectures to build the cross-modal retrieval model. Specifically, we include two image encoders, ResNet and Swin Transformer, representing two types of the most typical image encoder structures, CNN and Transformer. For text encoder, we consider two representative structures, RNN and Transformer, and include three models in total: GRU, vanilla Transformer, and BERT.

\label{sec:disc_encoder}
\begin{figure}
    \centering
    \includegraphics[width = 0.95\linewidth]{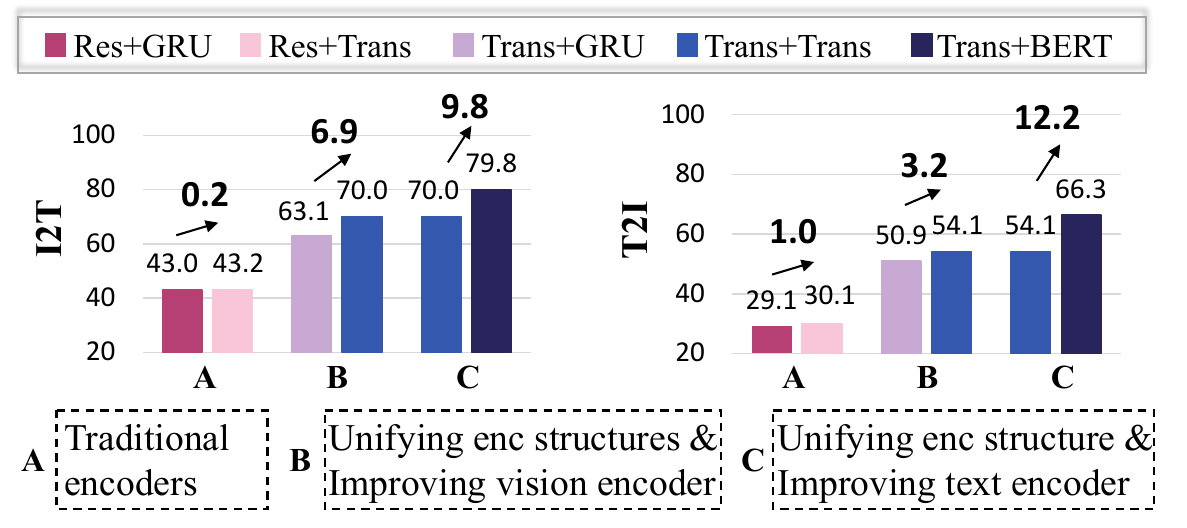}
    \caption{Performance (Recall@1) of models with different encoder combos and pairwise comparisons on Flickr30K.}
    \label{fig:encoders}
\end{figure}

Figure~\ref{fig:encoders} illustrates the performance improvements (Recall@1) of the models combining different image and text encoders for two cross-modal retrieval tasks on Flickr30K. From the comparison within pair \textit{A} we can first observe two models, \textit{i.e.}, ResNet+GRU and ResNet+Transformer achieve similar performance for both retrieval tasks, showing the worst performances, and the Transformer gains little improvement over GRU being text encoders when cooperated with ResNet. 
However, we observe great performance improvement (6.9 and 3.2 absolute improvements for I2T and T2I retrieval tasks) in pair \textit{B}, where the structures of two encoders are further changed to be the same, \myie{}, replacing the GRU by a vanilla Transformer for text encoder. This suggests the effectiveness of unifying the structures of image and text encoders. Moreover, when both image and text encoders employ the unified Transformer structure (pair \textit{C}), further advancing the text encoder with pre-trained and powerful BERT, can bring additional performance boost. This observation also supports the assertion of more powerful feature representation leads to better performance in previous parts.

\begin{figure*}
    \centering
    \includegraphics[width = \textwidth]{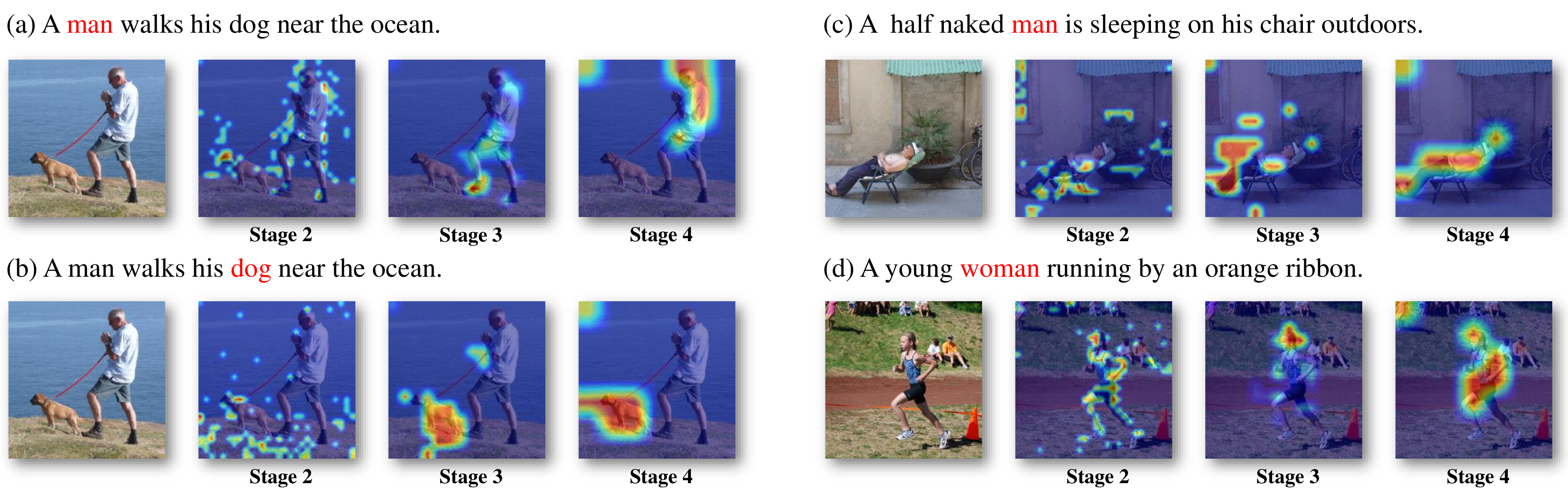}
    \caption{Illustrations of cross-attention map for text-to-image retrieval. We exhibit three-level attention maps for stage 2-4 of Swin Transformer to investigate the image-text associations of different levels.}
    
    \label{fig:hot}
\end{figure*}

\begin{table}[]
\centering
\caption{Performance comparison between HAT and its non-hierarchical variant on Flickr30K with the i-t cross-attention.} 
\setlength{\tabcolsep}{1.5mm}{
\begin{tabular}{c|ccc|ccc}
\hline
& \multicolumn{3}{c|}{Image-to-Text} &\multicolumn{3}{c}{Text-to-Image} \\ 
\cline{2-7} 
\multirow{-2}{*}{Model} & R@1 & R@5 & R@10 & R@1 & R@5 & R@10\\
\hline

w/o hierarchical & 79.8 & 95.9 & 97.5 & 66.3  & 88.6 & 93.3 \\ 
HAT(i-t) & 83.9 & 96.7 & 98.3  & 68.8& 90.5 & 94.6 \\ 
\hline
\end{tabular}
}
\label{tab:encoder}
\end{table}

\subsection{Analyses of Hierarchical Alignment}
To investigate the impact of our hierarchical alignment strategy, we conduct ablation for quantitative analysis and visualize the attention map of each layer for intuitive analysis. We first analyze the quantitative results, as illustrated in Table~\ref{tab:encoder}, the hierarchical alignment brings evident improvement for both image-to-text and text-to-image retrieval tasks. This means that our hierarchical alignment leveraging complementary multi-level features is able to enrich semantic associations between images and texts, and is beneficial to the overall retrieval goal.
To straightforwardly understand the learned feature at different levels, we further visualize the attention map in images for specific words of each level, as illustrated in Figure~\ref{fig:hot}. The shallow layer (Stage 2) tends to distribute the attention to many isolated dots or pixels around the target. While the attention map of middle layer (Stage 3) exhibits primary context-aware ability, and the deep layer (Stage 4) is able to capture the salient context with respect to the target word. 
The integration of multi-level could leverage the high-semantic context and low-semantic pixels to associate cross-modal information.
From the above observations, we can find that both the quantitative and qualitative results have verified that our hierarchical alignment strategy can capture more semantic context and achieve better performance. 

\subsection{Visualization of Image and Text Representations}
\begin{figure}
\begin{center}
\includegraphics[width = \linewidth]{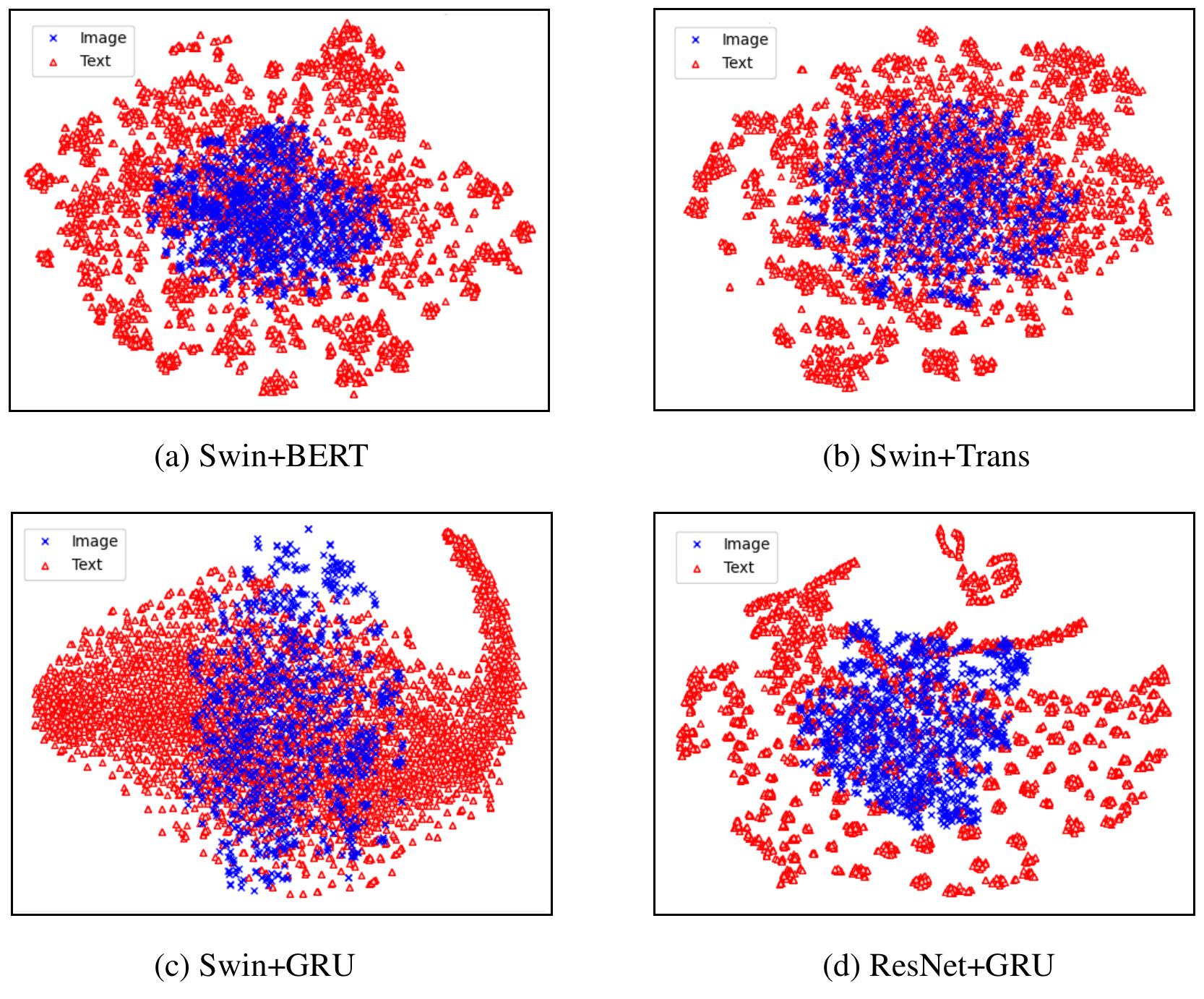}
\end{center}
\caption{
Distributions of learned image and text representations with different encoder combos. Both (a) and (b) employ Transformer architecture as image and text encoders, which show similar distributions for image and text representations. (c) and (d) show very different distribution patterns for image and text representations, due to the discrepancy between image and text encoder architectures.
}
\label{fig:tsne}
\end{figure}

In this part, we study the impacts on learned representation distribution with different encoder architectures. We utilize t-SNE~\cite{van2008visualizing} to visualize the distributions of image and text representations with Flickr30K test set with different architecture combos for the encoders.
As illustrated in Figure~\ref{fig:tsne}, we can observe that Swin+BERT (a) and Swin+Trans (b) employing Transformer architecture for both image and text encoder learn similar distributions for images and texts. While Swin+GRU (c) and ResNet+GRU (d) exhibit very different distribution patterns for image and text representations. These observations further support that the architecture discrepancy between encoders leads to different feature distribution for images and texts. Besides, the unified structure of two-stream encoders, Transformer architecture in specific, enables the encoders learn more compatible representation distributions for images and texts, and makes them align better in the common semantic space, resulting in better retrieval performance.

\section{Conclusions}
In this work, we studied the problem of architecture discrepancy between two-stream encoders in cross-modal retrieval. To address this issue, we proposed a novel framework, namely hierarchical alignment Transformers (HAT), which employed two-stream Transformers, \myeg{}, Swin Transformer and BERT, to unify the architectures of image and text encoder, and enriched the cross-modal associations by hierarchical alignment. The extensive experiments on two commonly-used datasets, Flickr30K and MSCOCO, demonstrated the superiority of our proposed HAT. Besides, further analyses verified the effectiveness of unifying encoder architectures of images and texts with Transformer, as well as the valuable ability of hierarchical alignment strategy.

\begin{acks}
This research/project is supported by the National Research Foundation, Singapore under its Industry Alignment Fund – Pre-positioning (IAF-PP) Funding Initiative. Any opinions, findings and conclusions or recommendations expressed in this material are those of the author(s) and do not reflect the views of National Research Foundation, Singapore. This work was also partially supported by the National Natural Science Foundation of China
under grant 62102070, 62220106008, and U20B2063, and partially supported by Sichuan Science and Technology Program under grant 2023NSFSC1392.
\end{acks}

\bibliographystyle{ACM-Reference-Format}
\bibliography{hat}


\end{document}